\documentclass[10pt,twocolumn,letterpaper]{article}

\usepackage[arxiv]{cvpr}      

\usepackage{graphicx}
\usepackage{amsmath}
\usepackage{amssymb}
\usepackage{booktabs}

\usepackage[pagebackref,breaklinks,colorlinks]{hyperref}

\usepackage[capitalize]{cleveref}
\crefname{section}{Sec.}{Secs.}
\Crefname{section}{Section}{Sections}
\Crefname{table}{Table}{Tables}
\crefname{table}{Tab.}{Tabs.}

\def\cvprPaperID{8273} 
\def\confName{CVPR}
\def\confYear{2023}

\begin{document}

\title{Divide and Compose with Score Based Generative Models}

\author{Sandesh Ghimire\\
{\tt\small drsandeshghimire@gmail.com}
\and
Armand Comas \\
{\tt\small comasmassague.a@northeastern.edu}
\and
Davin Hill\\
{\tt\small dhill@ece.neu.edu}
\and
Aria Masoomi \\
{\tt\small masoomi.a@northeastern.edu}
\and
Octavia Camps*  \\
{\tt\small camps@coe.neu.edu}
\and
Jennifer Dy * \\
{\tt\small jdy@ece.neu.edu}\\
Department of Electrical and Computer Engineering, Northeastern University\\
Boston, MA, USA
}
\maketitle
\begin{abstract}
   While score based generative models, or diffusion models, have found success in image synthesis, they are often coupled with text data or image label to be able to manipulate and conditionally generate images. Even though manipulation of images by changing the text prompt is possible, our understanding of the text embedding and our ability to modify it to edit images is quite limited. Towards the direction of having more control over image manipulation and conditional generation, we propose to learn image components in an unsupervised manner so that we can compose those components to generate and manipulate images in informed manner. Taking inspiration from energy based models, we interpret different score components as the gradient of different energy functions. We show how score based learning allows us to learn interesting components and we can visualize them through generation. We also show how this novel decomposition allows us to compose, generate and modify images in interesting ways akin to dreaming. We make our code available at {\textcolor{red}{https://github.com/sandeshgh/Score-based-disentanglement}}
\end{abstract}
\section{Introduction}
\label{sec:intro}
Diffusion based \cite{Ho_ddpm} or score based \cite{song2021scorebased} generative models are a new class of generative models based on the idea of reversing the image corruption process to generate realistic images from noise. These approaches have recently become quite successful, not only in synthesizing realistic and diverse images \cite{diffusion_beats_gan}, but also in obtaining better data likelihood \cite{diffusion_variational}.  Numerous works have applied score based generative models in text-based image generation \cite{latent_diffusion, dalle2}, inpainting \cite{latent_diffusion}, editing \cite{sdedit_meng} etc. Recently developed models like DALL-E\cite{dalle2} and Latent Diffusion \cite{latent_diffusion} using diffusion models have been reported to generate realistic and diverse images with wild imagination ability. 

Most of the works that conditionally generate images using diffusion/score model train models in a supervised manner conditioned either on actual class labels or embedding of paired text \cite{dalle2}. Supervised conditional generation could be either guided using the gradient of pretrained classifier \cite{diffusion_beats_gan} obtained from supervised learning or could also be classifier free \cite{classifier_free_diffusion}. Building upon the works of text based conditional generation, some works have also tried to edit or manipulate images \cite{avrahami2022blended}. While these methods show that with labels, learning conditional score model is quite effective, there is one fundamental problem with the present conditional generation: we do not have control over what the model generates. Suppose we generate an image based on text. The image looks okay, but it's not quite what we want. How can we change it to match our expectations? Do we have control? Do we have interpretable understanding of the conditioning? No!

This topic is not unexplored in the context of traditional generative models, like VAEs and GANs. In fact, they have been extensively explored as disentanglement \cite{mathieu2019disentangling,tcvae_chen2018} in the autoencoding setup and GAN inversion \cite{bau2020semantic,collins2020editing} in GANs. Several works in GAN inversion \cite{gan_inversion} try to find latent representations corresponding to an image and then manipulate the representation in the latent space to edit and manipulate the image. In case of score based models, however, there is little understanding of the latent factors in terms of generation of images. Models that use text based conditional generations are opaque and our only way to manipulate the image is through the generation of another text. 

To bridge this gap, we are interested in learning interpretable factors in a score based generative models, which could be later used to manipulate and edit images as in the case of GAN Inversion. The first plausible step to learn such factors is an autoencoding type approach where we first learn different image components and then recompose image out of those components. Unfortunately, the theoretical formulation of such diffusion autoencoder is currently unclear. Our first contribution is to cement the theoretical foundation of diffusion autoencoder through likelihood based formulation (see section \ref{sec:likelihood}). From the implementation perspective, we did find the diffusion autoencoder (DiffAE) implementation due to Preechakul et al. \cite{preechakul2021diffusion} to work well and have built upon their implementation.

While we agree with DiffAE \cite{preechakul2021diffusion} on the autoencoding setup, we take a very different approach to autoencoding by decomposing an image into different score components. We believe this is arguably better suited for the score models since the score functions are the main building components of score models. Therefore, we would like to decompose an image into different score components and try to understand their contribution in image generation. We take inspiration from the energy based models \cite{gen_convnet, flow_contrastive_ebm, comet}. Imagine that the probability density of image is given by product of exponential distributions of the form:
\begin{align}
\label{product_dist}
    p_0(x) = \prod p_0^{(i)}(x) = \frac{e^{-\sum_i E^{(i)}(x)}}{Z}
\end{align}
where $E^{(i)}(x)$ represents energy functions and $Z$ is the normalization constant, also known as partition function. Taking logarithm of eq.(\ref{product_dist}), we can conclude that modeling score as summation of several components can be interpreted as learning different energy components, \textit{i.e.} $s = \nabla_x \log p_0(x) = \sum_i \nabla_x (-E^{(i)}(x)) $. Based on this intuition, we imagine that we should be able to decompose the score function into different components and train the score based generative model. We ask, could we decompose an image into several interpretable components in unsupervised way and recombine them to generate new images, akin to dreaming? In this paper, we follow this abstract idea to generate interesting images by dividing image and recomposing components.

We perform several experiments to illustrate the score components learned in unsupervised manner and what we can achieve through their composition and manipulation. To interpret the factors captured by score components we generate images from individual components. Some components capture human interpretable attributes like shape and color, while others are not as they capture complex texture/features in images. We also modify images by manipulating individual components and interpolating them with unconditional score which results in interesting manipulation of images and diverse generation. We discuss how our experiments elicit new perspective on interpretability and disentanglement of images.

\section{Related Works}
Diffusion generative models based on denoising idea was first proposed by Ho et al. \cite{Ho_ddpm} and Sohl-Dickstein et al. \cite{sohl2015deep}.
From a different perspective, Song et al. \cite{song_datagradient} showed that we can generate images by estimating the score, i.e. gradient of data loglikelihood. These two perspectives were later reconciled by Song et al. \cite{song2021scorebased}. They showed that the forward diffusion and reverse generative models are both stochastic processes in continuous time guided by stochastic differential equations. This work unifies the diffusion perspective with the score based perspective. The score based generative model utilizes the denoising and implicit score matching ideas \cite{hyvarinen_score, vincent2011connection} to develop a computationally cheap way to estimate the score function at different time instants.

Score based generative model research has seen several new directions. Some works have tried to decrease the generation time with fast differential equation solvers \cite{fast_solver}, while others have tried to analytically estimate reverse time variance to improve image quality \cite{dpm_analytic}. Others have tried to improve the log likelihood of the score based generative models \cite{diffusion_variational}. Some theoretical works have derived the loss function from likelihood optimization perspective \cite{likelihood_chen2022,loglikelihood_song, likelihood_huang2021}. Other theoretical works have solved the stochastic differential equation by solving Schrodinger Bridge problem \cite{Bortoli2021diffusion}.

There are several applications of score based models like text based image generation \cite{dalle2,latent_diffusion}, image editing \cite{sdedit_meng} and adversarial purification \cite{adversarial_puri_diffusion, adversarial_puri}.

\section{Background}
\subsection{Denoising Diffusion Probabilistic Model}
Sohl-Dickstein et al. \cite{sohl2015deep} and Ho et al. \cite{Ho_ddpm} proposed to design a generative model, called Denoising diffusion probabilistic model (DDPM) from a Bayesian perspective. Imagine we sample an image from the data distribution, ${x}_0 \sim p_0$. Consider the data corruption sequence where we incrementally add Gaussian noise to the image  until it turns into complete noise. This forms a Markov chain and the joint distribution of the forward series is given by $p_0(x_0)\prod_{t=1}^{t=T} p_{t-1, t}(x_t)|p_{t-1, t}(x_{t-1})$. Then, a reverse Markov chain is considered were $p_{\theta}(x_{t-1}|x_t)$ is conditionally Gaussian such that the reverse process joint distribution is given by $p_{\theta}(x_T)\prod_{t=T}^{t=1} p_{\theta}(x_{t-1}|x_t)$. The DDPM algorithm optimizes the evidence lower bound of the data likelihood such that when optimization is complete, the reverse joint distribution coincides the forward joint distribution. Since the reverse conditional distributions are parameterized by the error functions $\epsilon_{\theta}$, the algorithm, essentially, boils down to optimizing the $\epsilon_{\theta}$. One key trick proposed in DDPM is that the complex loss obtained from the ELBO can be neatly expressed as an extremely simple loss function as follows:
\begin{align}
\label{ddpm_loss}
    \mathcal{L}_{\theta}^{ddpm} =  \mathbb{E}_{t, x_0,\epsilon } \big \{ \lambda_t ||\epsilon_{\theta}(x_t, t)
   -\epsilon||^2 \big \}
\end{align}
where, $x_t = \sqrt{\alpha_t}x_0+\sqrt{1-\alpha_t}\epsilon$ is a sample from distribution $p_t(x|x_0) = \mathcal{N}(x;\sqrt{\alpha_t}x_0,1-\alpha_t)$, the marginal distribution at time $t$ and $\epsilon$ is a random vector from an isotropic Gaussian distribution. Note that as we incrementally add noise to $x_0$, the marginal distribution $p_t$ at time $t$ can be expressed as a Gaussian distribution conditioned on $x_0$. The mean at time $t$ has diluted by a factor of $\sqrt{\alpha_t}$ and variance is $1-\alpha_t$. Also, the $\lambda_t$ in eq.(\ref{ddpm_loss}) is a function of $\alpha_t$s and noise added at each time instant.
Once we have the optimized network, $\epsilon_{\theta}$, the image generation process is nothing but following the reverse conditional distribution $p_{\theta}(x_{t-1}|x_t)$ starting from the isotropic Gaussian noise $p_{\theta}(x_T)$.

\subsection{Score Based Generative Model}
Song et al. \cite{song_datagradient} proposed a different generative model by estimating the gradient of data distribution, called the score function, whose sampling looked similar to that of DDPM. Even though the score based generative model seemed similar to diffusion model, DDPM, the connection was unclear until another paper due to Song et al. \cite{song2021scorebased}, which showed that there is a deeper connection between the two networks: the error network in DDPM, $\epsilon_{\theta}$ is same as the score function $s_{\theta}$ in score generative model. They establish that DDPM essentially performs score matching \cite{hyvarinen_score, vincent2011connection}. They further develop this line of argument showing that the continuous version of score based generative model can be obtained by reversing a stochastic differential equation (SDE). Specifically, they generalize the forward process of adding noise to a continuous setting where the forward process takes the form of a stochastic differential equation (SDE). Similarly, the reverse process of starting from the isotropic Gaussian and incrementally removing noise can also be shown to be a stochastic different equation continuous in time:
\begin{align}
    \text{FOR}:dx &= f(x,t)dt +g_t dw\\
    \text{REV}:dx &= [f(x,t) - g_t^2 \nabla_x \log p_t(x)] dt + g_t d\bar{w}
\end{align}
Luckily, the reverse stochastic differential equation only needs the score function at each timestamp $t$ in addition to other functions $f, g$ from the forward equation. Using the same score matching idea, the score function can be first trained with the following loss function
\begin{align}
\label{sde_score}
    \mathcal{L}_{\theta} =  \mathbb{E}_t \big \{ \lambda_t \mathbb{E}_{x_0} \mathbb{E}_{x_t| x_0}[||s_{\theta}(x_t, t)
   -\nabla_{x_t}\log p_{0,t}(x_t|x_0)||^2] \big \}
\end{align}

Note that this loss is same as eq.(\ref{ddpm_loss}) once we use the fact that $p_t$ is conditionally Gaussian. Once the score function is learnt, the generative model is given by replacing $s = \nabla_x \log p$ in the reverse time stochastic differential equation:
\begin{align}
\label{reverse_sde}
    dx = [f(x,t) - g_t^2 s_{\theta}(x,t)] dt + g_t d\bar{w}
\end{align}
In practice, we need to discretize eq. (\ref{reverse_sde}) to obtain a generative model. We achieve this through Euler-Maruyama discretization as suggested in \cite{song2021scorebased}:
\begin{align}
\label{euler_maruyama}
    x(t-\delta t) = [f(x,t) - g_t^2 s_{\theta}(x,t)] (-\delta t) + g_t \sqrt{\delta t}z
\end{align}
where $z \sim \mathcal{N}(0,I)$ is a random sample from standard Gaussian distribution.
\subsection{Conditional Score based Models}
Following the unconditional generative models, a few conditional generative models have been developed. Most of these models, however, require some form of supervision regarding the group on which to condition the generation. For example, Dhariwal et al. \cite{diffusion_beats_gan} developed conditional generative models based on class label by modifying score functions with extra term representing the gradient of log of classifier likelihood. These methods are known as classifier guided conditional generative models. Later, classifier-free models \cite{classifier_free_diffusion} were also proposed who eschewed the idea of training a classifier altogether. Nevertheless, they also need class label to train.

Towards the direction of unsupervised learning, Diffusion autoencoder (DiffAE) \cite{preechakul2021diffusion} learns the conditional distribution based on latent vector obtained from the encoder, which is in the line of our work. Our work differs from theirs in the fundamental notion of what represents the component. In our model, we encode different score components from an image, where each component sort of captures one concept (energy function). Later we generate images by recombining these components.
\section{Method}
Existing score based generative models train a common unconditional score function $s_{\theta}(x)$ such that the reverse SDE yields samples from the whole distribution $p_0(x)$. We are interested in learning different energy components in each image. Before we can decompose different energy components in each image, we need to be able to design a conditional score function which can reverse the forward diffusion process and converge to a single image rather than the whole distribution. More precisely, we want to modulate the score function such that the reverse SDE yields the dirac-delta distribution concentrated on a single image, say $x_{\zeta}$. It is unclear how to do that from the existing works. Note that the loss function in eq.(\ref{sde_score}) is an expectation across all images $x_0$ sampled from the data distribution $p_0$, and thus $s_{\theta}$ is common to all data. To design data-specific score function such that reverse SDE converges to a dirac delta distribution around image, we adopt the log likelihood formulation of score based models \cite{likelihood_huang2021, likelihood_chen2022, loglikelihood_song}.





\subsection{Likelihood Based Formulation}
\label{sec:likelihood}
We can derive a likelihood formulation to train the score based generative model based on Feynman-Kac Theorem \cite{karatzas1991brownian}, as shown in \cite{likelihood_huang2021, likelihood_chen2022, loglikelihood_song}.
We start from the likelihood formulation of training score based generative model. 
\begin{align}
\label{likelihood}
\nonumber
   &\log p(x_0) \geq \mathcal{L}_{VLB}(x_0, \theta)= E_{p_T}[\log p_{\theta}(x_T)] \\
   & - \int_0^T  \mathbb{E}_{p_t}\bigg[\frac{1}{2}g_t^2 ||s_{\theta}(x_t, t)||^2 + \nabla_{x} (g_t^2 s_{\theta}(x_t, t) -f )\bigg] dt
\end{align}
Taking expectation, we arrive at,
\begin{align}
\label{expected_likelihood}
   \mathcal{L}_{EVLB}(\theta) = \mathbb{E}_{{x_0} \sim p_{0}}[\mathcal{L}_{VLB}(x_0, \theta)]
\end{align}
From eq.(\ref{expected_likelihood}), we can derive the same loss function as in eq.(\ref{sde_score}) by using a rough approximation of the integral with the discrete summation and equivalence between different score matching \cite{vincent2011connection, hyvarinen_score}(as shown in \cite{likelihood_huang2021}). Therefore, unconditional score estimation can be thought as a crude approximation of the expected likelihood maximization. Nevertheless, expressing the likelihood as the integral in eq.(\ref{likelihood}) is much more illuminating and powerful. Observe that eq.(\ref{expected_likelihood}) is obtained by taking expectation of the likelihood of individual data in eq.(\ref{likelihood}). It is this expectation which leads to learning a common unconditional score function. We can forego expectation to train score functions for each data point. For any data $x_{\zeta}$, we can directly optimize its likelihood by training a score function $s_{\theta, \zeta}$ to optimize the lower bound on the right hand side of eq.(\ref{likelihood}). Specifically, we train an encoder, \texttt{Enc} and score function, $s$ as follows: 
\begin{align}
\label{eq_encode}
&\zeta = \texttt{Enc}_{\theta}(x_{\zeta})\\
\nonumber
\label{eq_loglikeli}
&\log p(x_{\zeta}) \geq = E_{p_T}[\log p_{\theta}(x_T) | x_0 = x_{\zeta}] \\
   & - \int_0^T  \mathbb{E}_{p_t}\bigg[\frac{1}{2}g_t^2 ||s_{\theta}||^2 + \nabla_{x} (g_t^2 s_{\theta, \zeta} -f )| x_0 = x_{\zeta}\bigg] dt
\end{align}
Eq.(\ref{eq_encode}) and eq.(\ref{eq_loglikeli}) completes our autoencoder model that maximizes the lower bound on log likelihood of individual image, $x_{\zeta}$.

\subsection{Score Decomposition}
As motivated in the introduction (eq.(\ref{product_dist})), we want to decompose each score function into multiple components, which intuitively represent different energy components (negative of gradient of energy to be precise). We decompose each score function into $K$ components such that $s_{\theta, \zeta} = (s_{\theta, \zeta}^{(1)} +s_{\theta, \zeta}^{(2)} + ...+s_{\theta, \zeta}^{(K)})/K$. This decomposition requires $K$ different score functions. A computationally efficient way would be to share the weight of score functions but use different latent vectors for different components, which also gives a structure to the latent space. Therefore, we decompose the score function as:
\begin{align}
\label{score_sum}
    s_{\theta, \zeta} = (s_{\theta, \zeta^{(1)}} +s_{\theta, \zeta^{(2)}} + ...+s_{\theta, \zeta^{(K)}})/K
\end{align}
Note that now the burden of learning different energy components has been shifted to the latent vectors together with a shared conditional score function $s_{\theta, \zeta^{(k)}}$.

\subsection{Model and Training}
The encoder encodes each image $x_{\zeta}$ into $K$ latent vectors $\zeta^{(K)}$. The summed score function given by eq.(\ref{score_sum}) is used to maximize the lower bound on the log likelihood of each $x_{\zeta}$. We also approximate this integral with discrete summation and invoke the equivalence of implicit score matching \cite{hyvarinen_score} and denoising score matching \cite{vincent2011connection,song2021scorebased} to obtain an approximation \cite{likelihood_huang2021} as the following loss function:
\begin{align}
\label{our_loss}
    \mathcal{L}_{\theta, \zeta} =  \mathbb{E}_t \big \{ \lambda_t \mathbb{E}_{x_t| x_{\zeta}}[||s_{\theta, \zeta}(x_t, t)
   -\nabla_{x_t}\log p_{0,t}(x_t|x_{\zeta})||^2] \big \}
\end{align}   

We jointly train $K$ encoders and the score loss in eq.(\ref{our_loss}), where the score is given by eq.(\ref{score_sum}). To design a score function as a function of the the latent vector $\zeta$, we use the adaptive Group Norm (AdaGN) strategy as described in \cite{diffusion_beats_gan}. 
Once the model is trained, we generate image by first sampling noise $z$ from the standard Gaussian distribution and iteratively applying Euler-Maruyama discretization of the SDE as described by eq.(\ref{euler_maruyama}). 

We also experiment with generating samples by combining score components with unconditional score function similar to \cite{classifier_free_diffusion}. For this, we can separately train conditional and unconditional score functions. However, to make it computationally cheap, we pass a vector of ones $\textbf{1}$ to realize the unconditional score function $s_{u} = s_{\theta, \textbf{1}}$, similar to the trick used in \cite{classifier_free_diffusion}. We combine conditional and unconditional score functions with linear combination of coefficients as described in experiments.

\section{Results and Discussion}
\begin{figure}[t]
  \centering
   \includegraphics[width=0.99\linewidth]{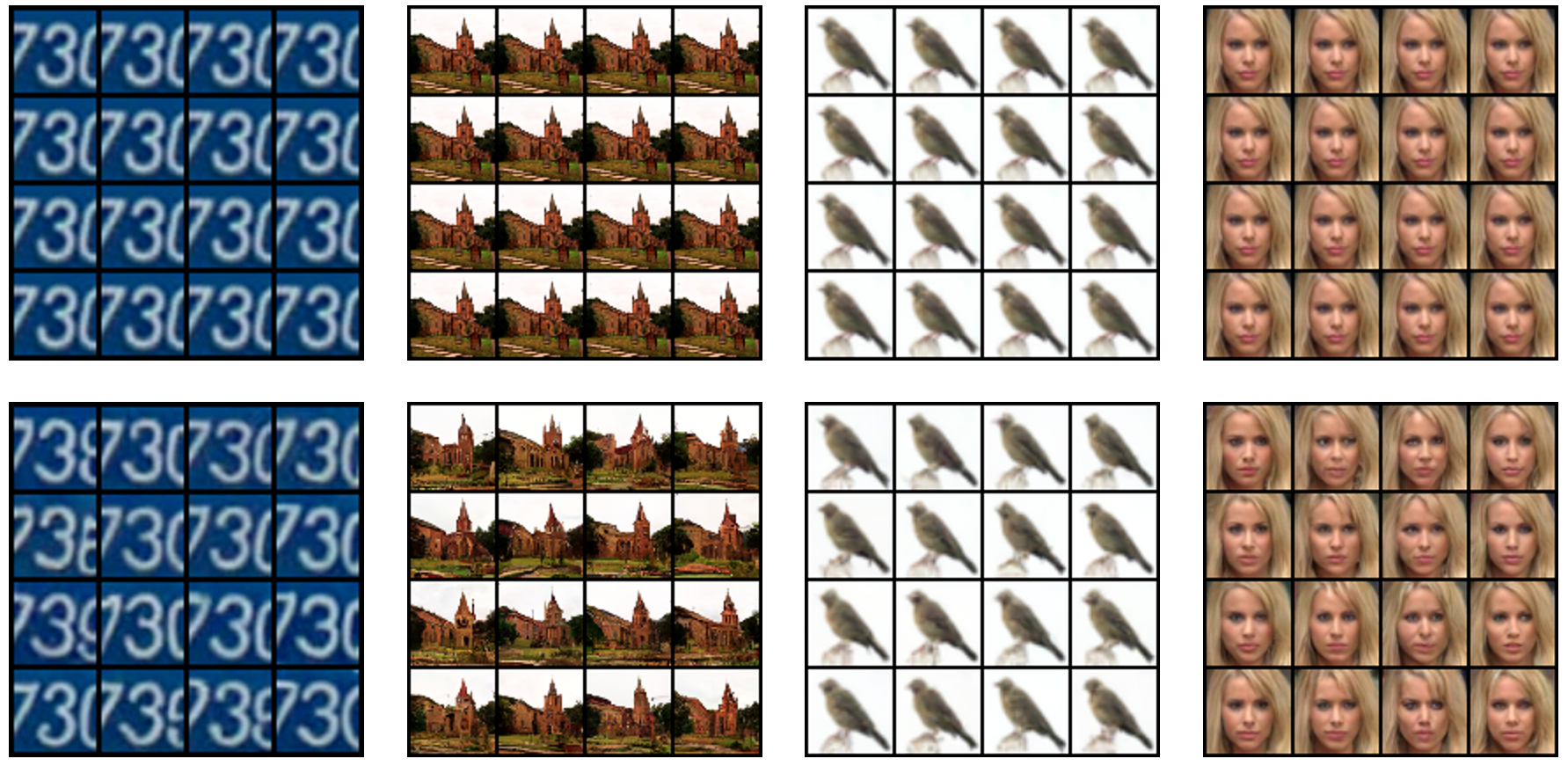}

   \caption{Using all score components results in faithful reconstruction, but with interesting variations}
   \label{fig:reconstruct}
\end{figure}
\subsection{Experimental Details}
\begin{figure*}[t]
  \centering
   \includegraphics[width=0.99\linewidth]{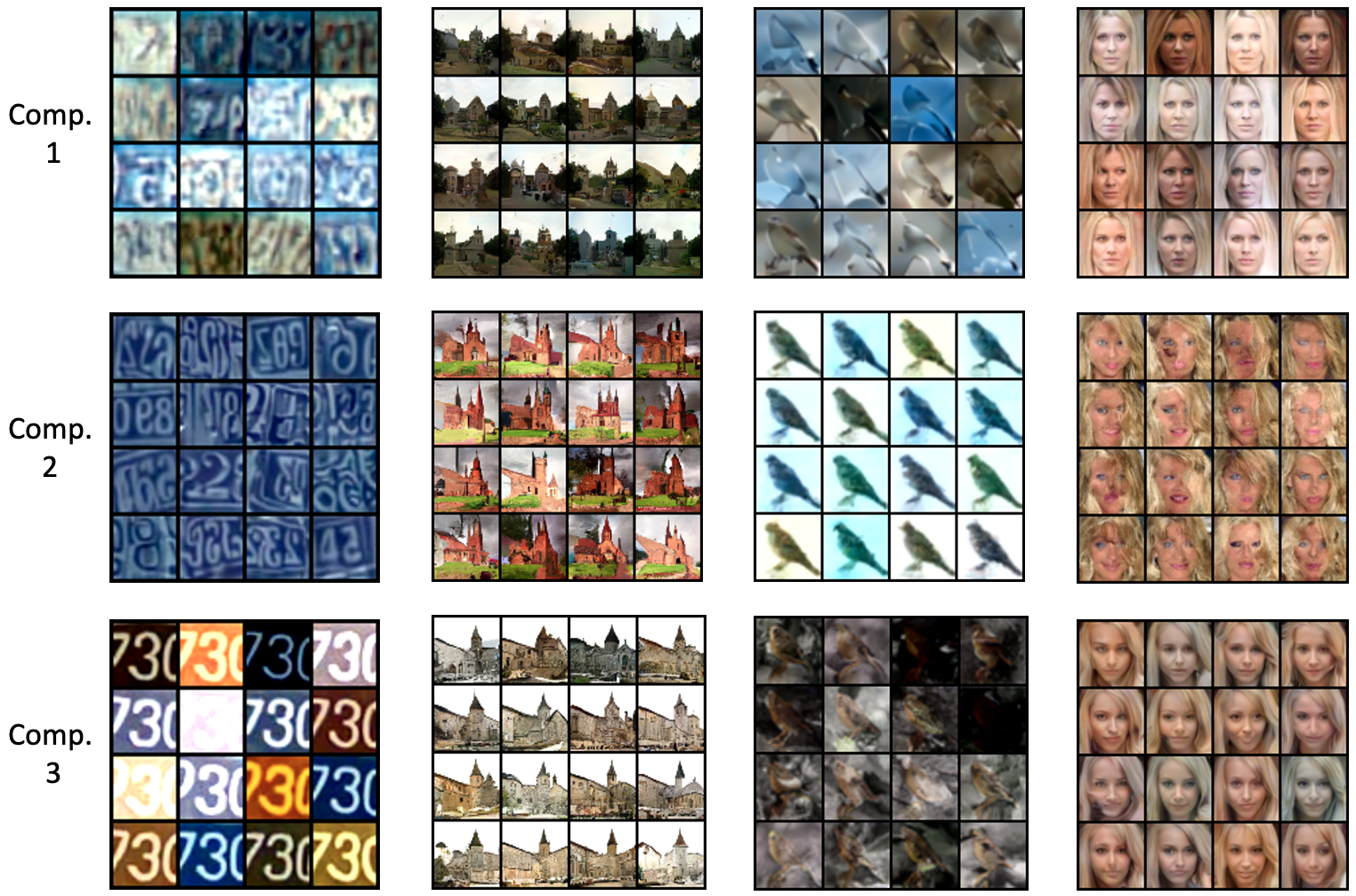}

   \caption{Visualizing different score components through generation as a way to interpret components. Images are generated by solving reverse SDE using single score component. Each row is a component and each column is different dataset.}
   \label{fig:three_comp}
\end{figure*}
The score network is the modified version of UNet architecture as described in Dhariwal et al.\cite{diffusion_beats_gan}. To condition on the latent vector obtained from encoder, we use the AdaGN as described in \cite{diffusion_beats_gan}, which is inspired from adaptive instance normalization (AdaIN) \cite{AdaIN}, but uses group normalization \cite{groupnorm} instead of instance norm. Our architecture is similar to DiffAE's adaptation of AdaGN, \textit{i.e.}
\begin{align}
\texttt{AdaGN}(h,t,\zeta) =  \zeta_s (t_s .\texttt{GroupNorm}(h)+t_b)    
\end{align}
where $h$ is normalized feature map at different layers, $t_s, t_b$ are obtained from the time embedding by applying MLP and $\zeta_s$ is affine transformation of the latent vector $\zeta$.

We burrow encoder from DiffAE \cite{preechakul2021diffusion}, i.e. the first part of UNet architecture used in score function. We experiment with two values of K: $K=\{3,5\}$. The encoder encodes K vectors of length $128$, which condition the score functions. 

Due to the computing constraints, we train for 150K iterations using Adam optimizer and batch size of 32. We experiment on four datasets in computer vision: 1) Celeb-A, 2) LSUN-outdoor\_church, 3) Cifar-10 and 4) SVHN. We use image size of $32 \times 32$ for Cifar-10 and SVHN and $48 \times 48$ for LSUN and Celeb-A, again due to resource constraints.


\subsection{Natural variation in reconstruction}
In Fig.(\ref{fig:reconstruct}), we show auto encoding of samples from four datasets. We input a batch of 16 images, all of which are the same. Figure shows that the reconstruction is very close to the ground truth, but, at the same time, there are natural variations. 
% ...

\subsection{Visualizing different components through generation}
\begin{figure*}[h!]
  \centering
   \includegraphics[width=0.99\linewidth]{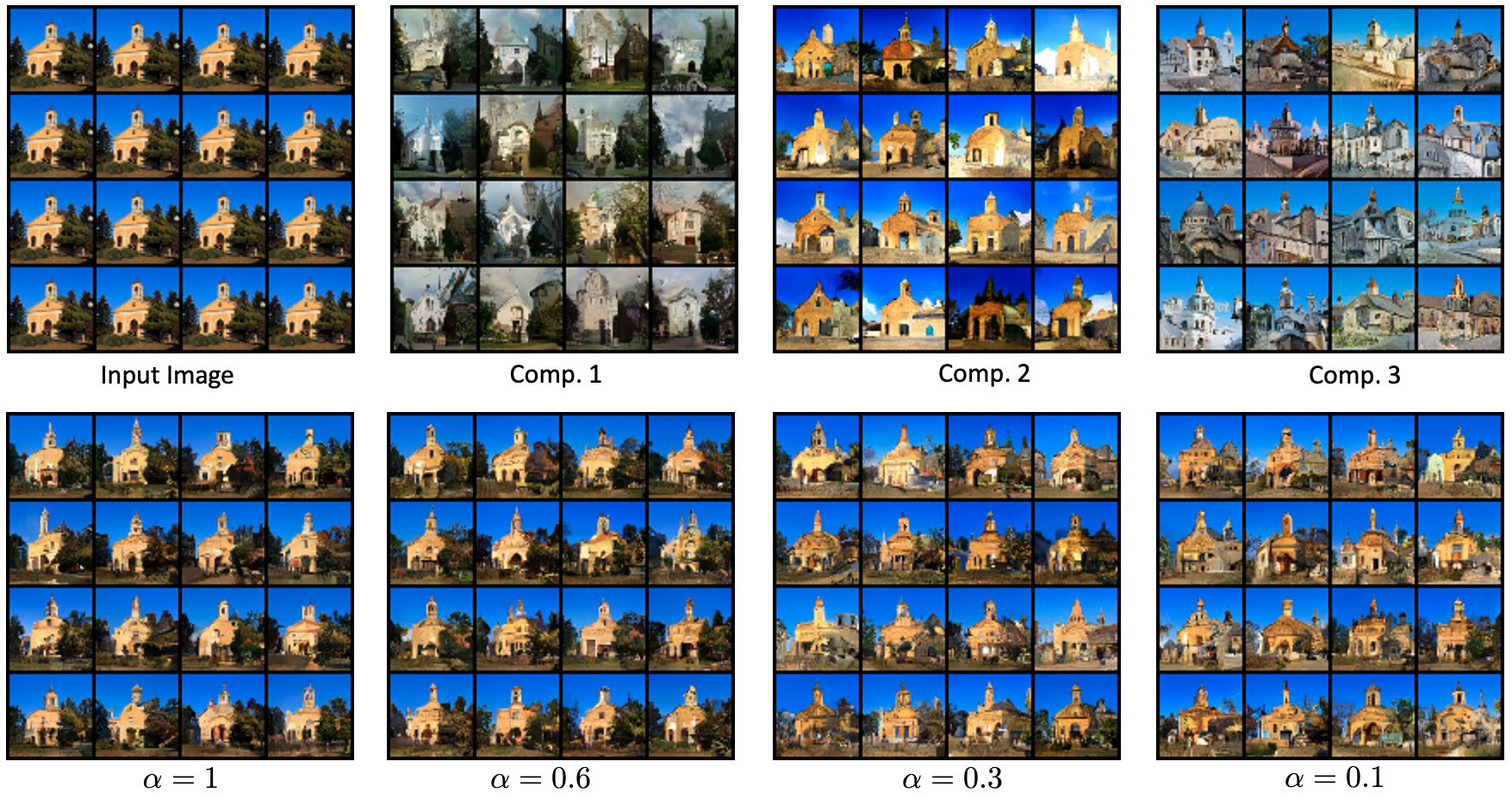}

   \caption{First row shows different components. Second row shows the effect of slowly diluting first component with unconditional score function. Decreasing $\alpha$ means weight of first component decreases and that of unconditional score function increases.}
   \label{fig:switch_1_lsun}
\end{figure*}
Visualization of different score components can be insightful about what each component is capturing from the image. It is unclear what is the best way to visualize score components which are essentially represented as matrices. We demonstrate that generating image samples using each component can be the best visualization strategy of each components. To be precise, if $s_{\theta, \zeta} = (s_{\theta, \zeta^{(1)}} + s_{\theta, \zeta^{(2)}} +s_{\theta, \zeta^{(3)}})/3$, then we take each component, say $s_{\theta, \zeta^{(k)}}$ and generate images using the Euler Maruyama discretization in eq. (\ref{euler_maruyama}). The samples generated using three components are plotted in Fig.\ref{fig:three_comp} where each row is a component. The input image of these components are the same images as in Fig.\ref{fig:reconstruct}. Note that even though input image and the latent vectors are the same, there are natural variations in samples form each component.

Several observations are in order. First, this method can result in human interpretable components in some cases, but not always. For example, the third component in the svhn image is clearly capturing digits, however, first and second components are capturing some abstract texture or lighting related information. Similarly, second component of the bird is clearly focusing on the color, but first one is capturing some abstract shape information. 

This score decomposition forces us to rethink the definition of components of an image. In classical works, disentanglement has been seen as some kind of statistical property of the distribution and methods were proposed to enforce such statistical pattern, for example, independence. Here, however, we see that components could capture interesting pattern that may or may not be independent or interpretable. We can certainly say that the three components are different, but at the same time capture something about the input image. Therefore, they are different disentangled components. Yet, they may or may not possess statistical independence property or complete human interpretability.

\begin{figure*}[t]
  \centering
   \includegraphics[width=1.0\linewidth]{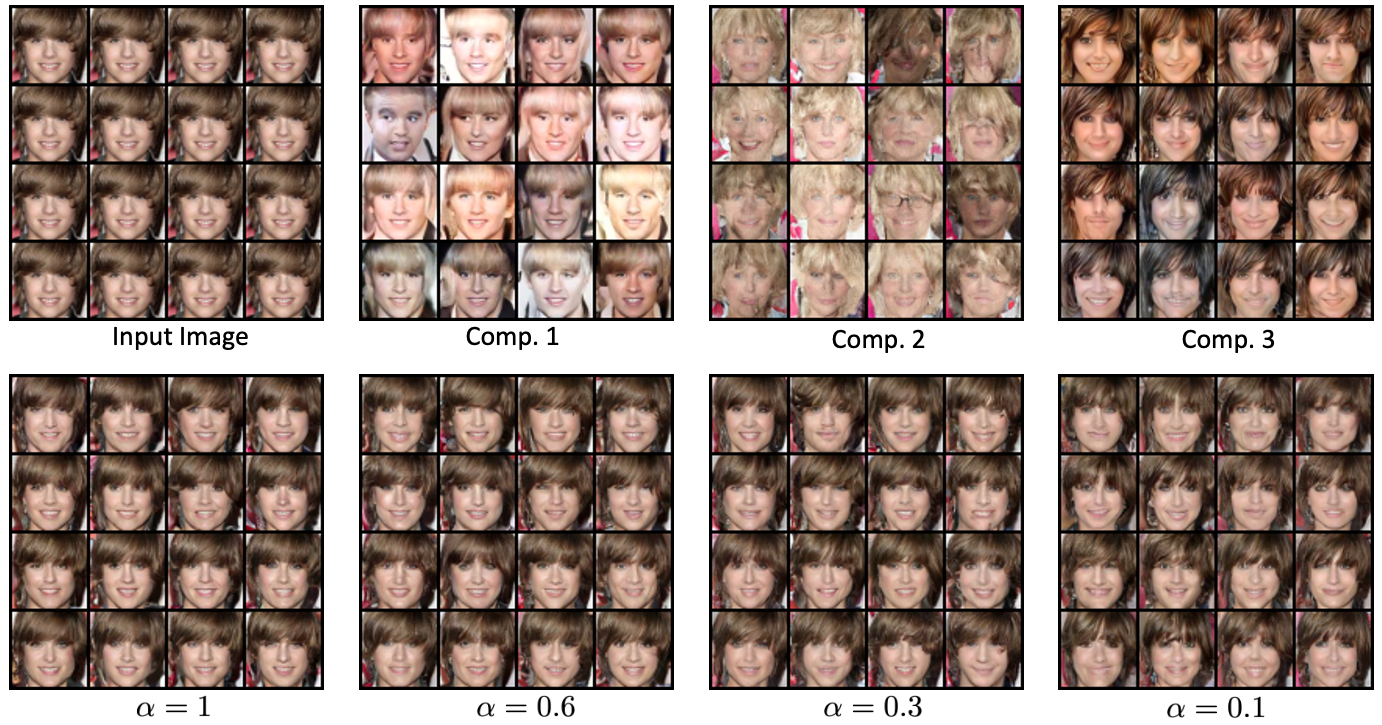}

   \caption{First component is diluted with unconditional score function}
   \label{fig:switch_1_celeba}
\end{figure*}
\begin{figure*}[h]
  \centering
   \includegraphics[width=0.99\linewidth]{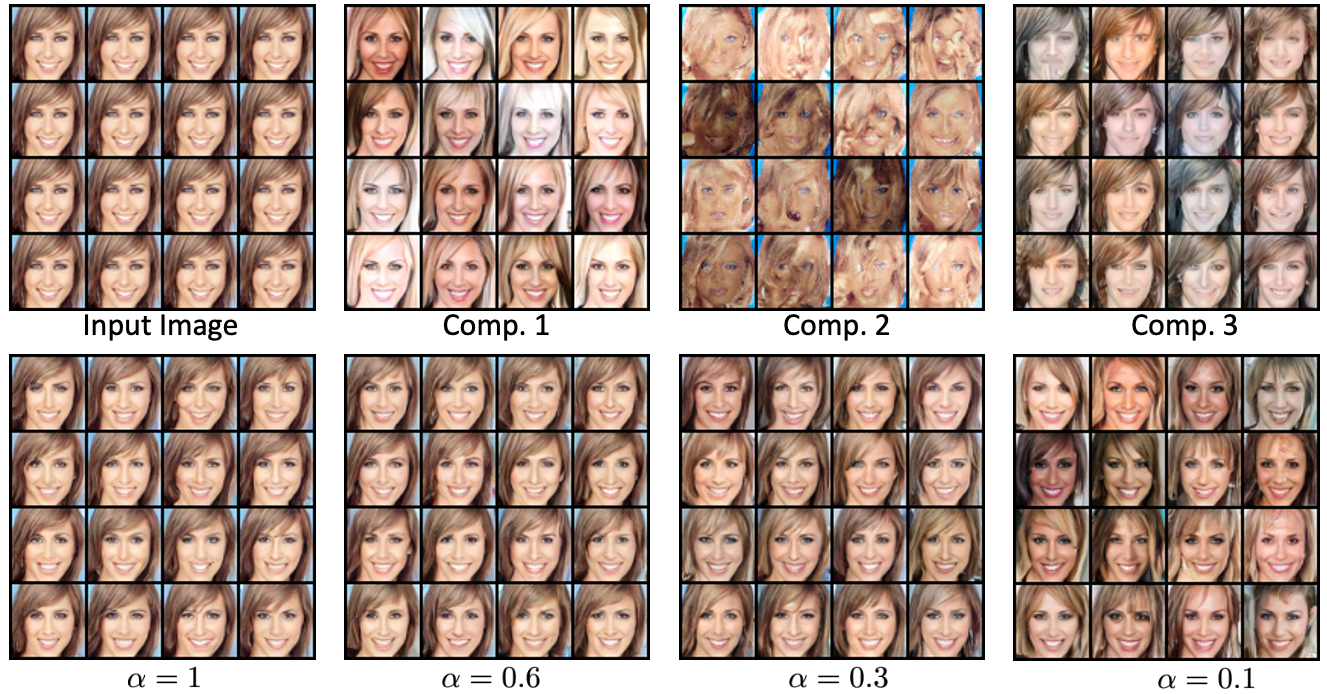}

   \caption{Second and third components are diluted with unconditional score function in second row.}
   \label{fig:switch_2_celeba}
\end{figure*}
\subsection{Manipulation of components}

We manipulate images using these learned components, with the help of an unconditional score function. In Fig.\ref{fig:switch_1_lsun}, we retain two components untouched while we linearly interpolate one component with the unconditional score function to generate images. That is, if $s^{(1)}, s^{(2)}, s^{(3)}$ are three components, we generate image with the following score function:
\begin{align}
 s_{\alpha}^{interp} = (\alpha *s^{(1)} + s^{(2)}+ s^{(3)})/3 +(1-\alpha)s_u/3 
\end{align}

In Fig.\ref{fig:switch_1_lsun}, we observe that the first component is associated with the background information and less in the church building. When this component is interpolated, we see that the information like yellow color (related to second component) and building architectures (related to third components) are preserved. As we go from $\alpha = 1$ to $\alpha =0.1$, we see that the image changes while preserving the yellow color of the church and with different building architectures, but it changes the tree and the background making it more diverse. At $\alpha = 0.1$, we see that a lot of tree has been removed and replaced with something else to create diverse background setting.

In Fig.\ref{fig:switch_1_celeba}, we observe similar effect. The first component is changed in second row while retaining second and third component. From the first row, it is clear that the first component is associated with smile. Hence, as we go from $\alpha = 1$ to $\alpha =0.1$, the sample images are more diverse in terms of smile. At $\alpha = 0.1$, we see a few instances where the mouth is even closed.

Similarly, in Fig.\ref{fig:switch_2_celeba}, we change second and third component while retaining the first one. Here also, the first component is associated with the smile while second and third are associated with shape features and hair. We fix the first component and interpolate second and third component as follows:
\begin{align}
 s_{\alpha}^{inter} = \frac{(\alpha *(s^{(1)} +  *s^{(2)})+ s^{(3)})}{3} +\frac{2(1-\alpha)s_u}{3}
\end{align}
In doing so, when we go from $\alpha =1$ to $\alpha = 0.1$, smile is preserved while other features change. Hence, we see image with a lot of diversity in terms of shape, color and hair while preserving smile as we go towards $\alpha = 0.1$.

\subsection{Tuning the weight of components}
\begin{figure*}[t]
  \centering
   \includegraphics[width=0.99\linewidth]{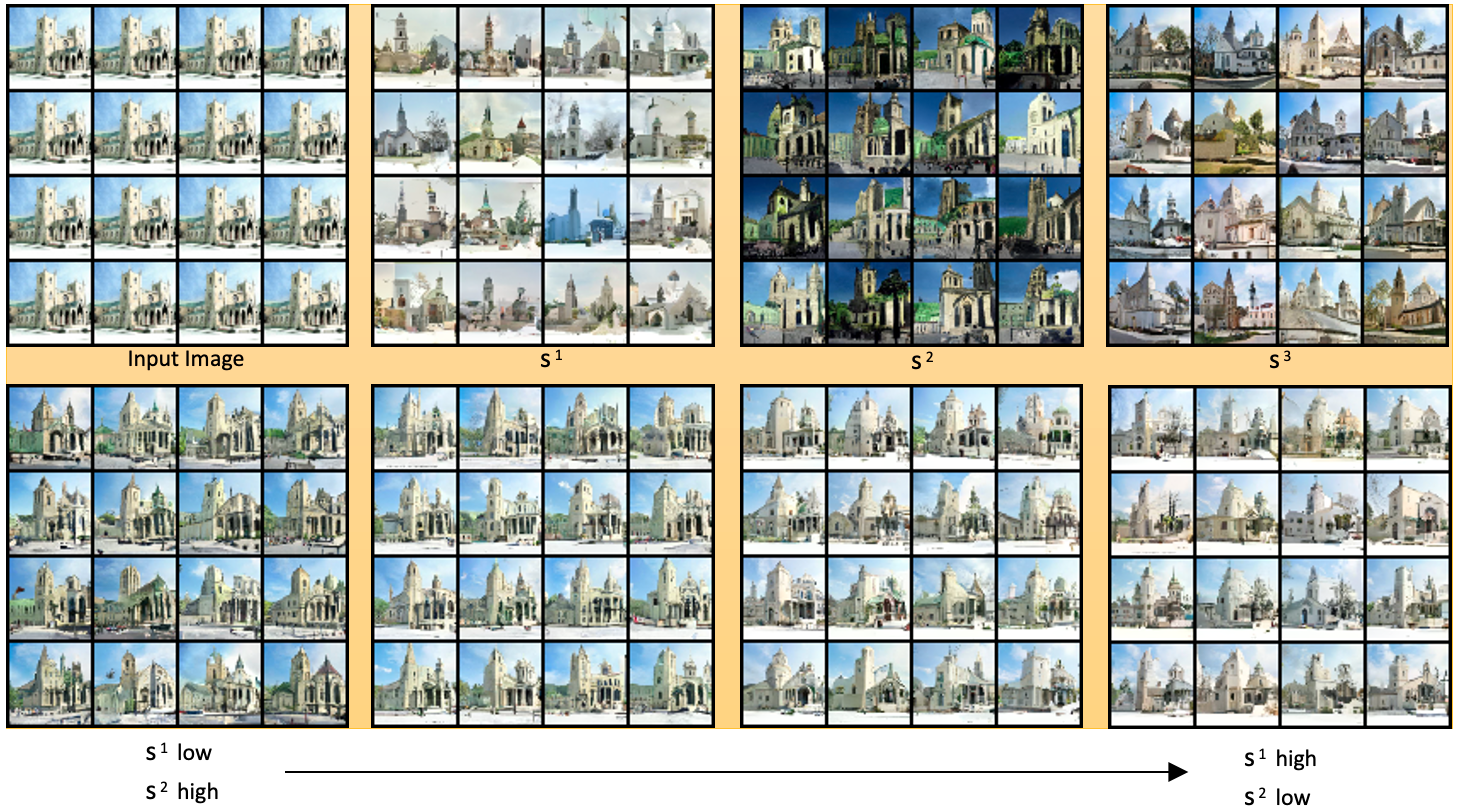}

   \caption{Second row: Changing the relative weight of first and second components while keeping the third component constant.}
   \label{fig:tune_lsun}
\end{figure*}

In Fig.\ref{fig:tune_lsun}, we tune the weight of two score components $s^{(1)}$ and $s^{(2)}$. When the $s^{(2)}$ is weighted heavily, the greenish hue and the corresponding shape of the church are dominating. As we moved more heavily towards $s^{(1)}$, we started seeing more of a mix between $s^{(1)}$ and $s^{3}$.
\subsection{Varying the number of components}

\begin{figure*}[h]
  \centering
   \includegraphics[width=0.99\linewidth]{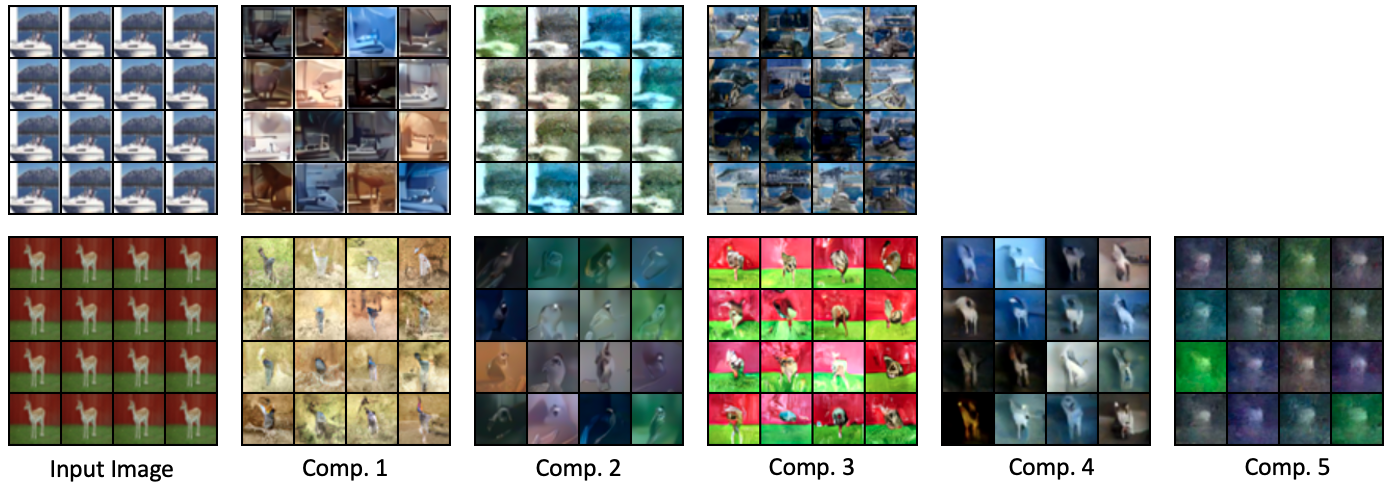}
   \caption{Visualizing three and five components in Cifar10 datase through image generation from components.}
   \label{fig:3vs5}
\end{figure*}
Fig.\ref{fig:3vs5} compares five components against three components learned in Cifar10 dataset. Components 4, 5 and 1 in five components are similar to the three in first row. However, Component 3 in second row is different and seems to capture the colored pattern with shape in the middle. 

\section{Conclusion}
Based on the insights of energy based models, we proposed to learn multiple score components while training a score based generative model. These components are interpretable and provide us more control to manipulate and edit images. In experiments, we discuss the interpretability and demonstrate our ability to edit and generate images through component manipulation. These score components also provide us new interpretable methods in score based generative models.

There are some rooms for improvement. Our ability to manipulate images is limited by the number of score components. Scaling this method to higher number of components without computational overhead could be an important future direction. Guiding components towards interpretations more amenable to humans could be another promising direction of future research.
\bigskip
{\small
\bibliographystyle{ieee_fullname}
\bibliography{egbib}
}

\end{document}